# Introducing Feature Attention Module on Convolutional Neural Network for Diabetic Retinopathy Detection

Susmita Ghosh
*Department of Computer Science and Engineering*
*Jadavpur University*
Kolkata, India
susmitaghoshju@gmail.com

Abhiroop Chatterjee
*Department of Computer Science and Engineering*
*Jadavpur University*
Kolkata, India
abhiroopchat1998@gmail.com

*Abstract*—Diabetic retinopathy (DR) is a leading cause of blindness among diabetic patients. Deep learning models have shown promising results in automating the detection of DR. In the present work, we propose a new methodology that integrates a feature attention module with a pretrained VGG19 convolutional neural network (CNN) for more accurate DR detection. Here, the pretrained net is fine-tuned with the proposed feature attention block. The proposed module aims to leverage the complementary information from various regions of fundus images to enhance the discriminative power of the CNN. The said feature attention module incorporates an attention mechanism which selectively highlights salient features from images and fuses them with the original input. The simultaneous learning of attention weights for the features and thereupon the combination of attention-modulated features within the feature attention block facilitates the network's ability to focus on relevant information while reducing the impact of noisy or irrelevant features. Performance of the proposed method has been evaluated on a widely used dataset for diabetic retinopathy classification e.g., the APTOS (Asia Pacific Tele-Ophthalmology Society) DR Dataset. Results are compared with/without attention module, as well as with other state-of-the-art approaches. Results confirm that the introduction of the fusion module (fusing of feature attention module with CNN) improves the accuracy of DR detection achieving an accuracy of 95.70%.

## I. INTRODUCTION

Diabetic retinopathy (DR) is a major cause of blindness among working-age adults worldwide. Early and accurate detection of DR is crucial for timely intervention and effective management of the disease. Researchers have used various techniques such as neural networks, fuzzy sets, nature inspired computing with an aim to enhance the accuracy of object tracking, object segmentation, object detection tasks making it relevant for computer vision applications [1-4]. Several researchers contributed in the field of medical image analysis for improved disease diagnosis and detection. In recent years, deep neural nets [5] have shown remarkable advancements in various computer vision tasks, including medical image analysis. In this article, we focus on DR detection using deep learning, with specific emphasis on the integration of a novel feature attention block with a pretrained VGG19. Detecting DR involves the analysis of retinal fundus images which can provide valuable insights into the progression of the disease.

Traditional approaches to DR detection relied on handcrafted features and shallow classifiers, which often struggled to capture the intricate patterns and subtle characteristics indicative of diabetic retinopathy. However, deep learning models (Fig. 1), with their ability to automatically learn features from raw data, demonstrated great potential in improving the accuracy of DR detection.

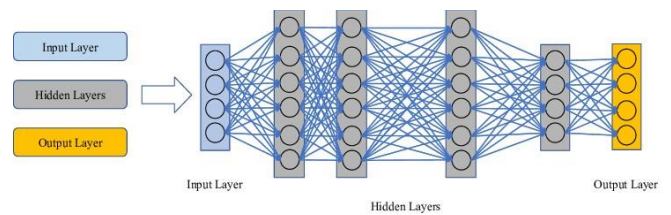

**Fig. 1:** General representation of a deep neural network [6]

In the present work, we propose a new feature attention block at a distinct position that enhances the discriminative power of the pretrained VGG19 architecture for DR detection. The attention block selectively amplifies the informative regions within the input image, allowing the network to focus on relevant features while suppressing irrelevant/noisy information. By incorporating this block after the VGG19 backbone, the model's ability to extract and emphasize on crucial features increases.

As mentioned, the primary objective of this article is to design a neural network model that could provide better DR detection accuracy. To do so, the performance of our proposed method is evaluated with those of the baseline VGG19 architecture without the feature attention block. We have considered the popularly used APTOS 2019 [7] Blindness Detection Challenge dataset. Results are compared in terms of various performance metrics. Results establish the superiority of the newly designed model in comparison to seven other state-of-the-art techniques.

The remainder of this paper is organized as follows: Section 2 provides a review of related works in the field of DR detection using deep learning. Section 3 presents the methodology, including a detailed description of the proposed feature attention block, its integration with the VGG19 architecture, and the transfer learning technique used. Section 4 discusses the experimental setup mentioning dataset used, evaluation metrics considered and details of parameters taken. Analysis of results has been put in Section 5. Finally, Section 6 concludes the paper.

By introducing a new feature attention block to enhance the performance of pretrained models on the APTOS dataset, our research contributes to the ongoing efforts in improving the early detection of diabetic retinopathy.

## II. RELATED RESEARCH

Gulshan et al. [8] and Abràmoff et al. [9] proposed CNN models for automated screening of diabetic retinopathy. They trained a deep neural network using a dataset of retinal fundus images. The CNN architecture consisted of multiple convolutional layers followed by max pooling layers to extract features from images. The extracted features were then passed through fully connected layers for classification. Quellec et al. [10] introduced a joint segmentation and classification approach for DR lesions. They used a model that combined CNNs with a conditional random field framework, performing segmentation and classification simultaneously.

Burlina et al. [11] focused on the classification of age-related macular degeneration (AMD) severity, a condition related to DR. They used a deep model called *DeepSeeNet*, which employed a CNN architecture trained on a large dataset of retinal images. Chaudhuri et al. [12] provided a comprehensive analysis of deep learning models for DR detection. They explored various CNN architectures, including VGGNet, Inception-v3, and ResNet. The authors compared the performance of these models and discussed their strengths and limitations. Gargeya and Leng [13] presented a review of deep learning-based approaches for DR screening. They discussed different CNN architectures including AlexNet, GoogLeNet, ResNet.

## III. METHODOLOGY

In the present work, we propose a neural network model for DR detection using deep learning specifically focusing on the integration of a new feature attention block with a pretrained VGG19 architecture. Block diagram of the proposed method is shown in Fig. 2.

As shown in Fig. 2, the first layer applied is the Weighted Global Average Pooling which takes the features obtained through the pretrained VGG19 as input. It computes the weighted average of the feature maps along the spatial dimensions, resulting in a tensor with reduced spatial dimension *(batch_size, 1, 1, channels)*. This layer focuses on capturing the importance of each channel based on weighted average.

Thereafter, two dense layers are employed. The first one uses a rectified linear unit (ReLU) activation function reducing the number of channels to *chan_dim/ratio*. This layer introduces non-linearity and compresses the channel dimension. The resulting tensor has a shape *(batch_size, 1, 1, chan_dim/ratio)*. The second dense layer utilizes a sigmoid activation function to produce an output tensor with the same shape as that of the first one. This layer focuses on determining the channel-wise importance through a sigmoidal activation that assigns attention weights to each channel. These attention weights (importance) are applied to feature maps.

To incorporate the computed attention weights, the input feature tensor is multiplied element-wise with the output tensor obtained from the second dense layer. The resulting tensor retains the original spatial dimensions of the input feature while emphasizing important channel activations. Its shape is *(batch_size, height, width, chan_dim)*.

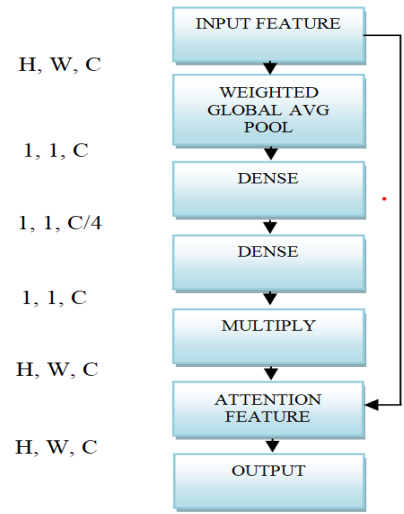

**Fig. 2:** Block diagram of the feature attention module. The output dimensions are shown on the left side.

To preserve original information and facilitate gradient flow, a skip connection is incorporated to add the attention feature tensor obtained from the element-wise multiplication to the input feature tensor. This process forms a residual connection, ensuring that the original information is retained while incorporating channel attention. The output tensor has the same shape as that of the input feature, i.e., *(batch_size, height, width, chan_dim)*.

**Feature Attention Mechanism:**

The new feature attention block, along with skip connection, leverages the power of weighted global average pooling to further capture the importance of each channel in a feature map while preserving valuable channel-wise details. By calculating the weighted average activation of each channel across spatial dimensions, the technique effectively condenses spatial information.

The skip connection is vital for information flow and preserves important details throughout the network. This is achieved through the addition of the attention feature tensor to the original input feature map.

By applying subsequent operations involving dense layers with ReLU and sigmoid activations, attention weights are generated to provide a measure of relevance/importance of each channel. As mentioned, these weights are then applied to the feature map using element-wise multiplication, dynamically amplifying the contribution of informative channels while attenuating the less important ones. This selective emphasis on important channels enables the model to focus on relevant features and extract discriminative information effectively.

The proposed methodology enhances the performance of DR detection through integration of transfer learning and the attention module. Transfer learning involves utilizing pretrained VGG19 net, (trained on ImageNet dataset), as feature extractor. This model captures powerful visual representations that are generalizable to various tasks. The attention module is augmented after the VGG19 layers (Fig. 3), selectively emphasizing important channels in the extracted features. Overall, the model benefits from both the learned representations and the specialized attention mechanism. The block diagram of the proposed methodology, augmented with feature attention module is shown in Fig. 3. Relevant working details are described below.

Let *I* be the input tensor of dimension *H×W×C*, where *H* and *W*, respectively, represent the height and width of the channel (feature map), and *C* represents the number of input channels. The average pooling operation (*a_p*) is defined as:

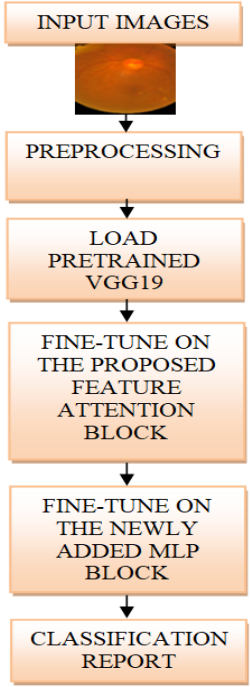

**Fig.3:** Block diagram of the proposed methodology (augmented with feature attention block)

$$a\_p = \left(\frac{1}{H \times W}\right) \sum_{x=1}^{H} \sum_{y=1}^{W} i\_f(i,j,k) \quad (1)$$

where, *i* ranges from *1* to *batch_size*, *j* ranges from *1* to height and *k* ranges from 1 to width of the input image, *i_f* represents *input_feature*. The resulting tensor *a_p* has a shape *(batch_size, 1, 1, chan_dim)*. *batch_size* represents the number of images present in each batch during training.

The output of the first dense layer, *fc1*, is written as:

$$fc1(i,1,1,k) = ReLU(W1(e,d).a\_p(i,1,1,d) + b1(k)). \quad (2)$$

This reduces the dimensionality of the input by multiplying the average-pooled features *avg_pool(i,1,1,d)* with the corresponding weights *W1(k,d)*, summing them up, and adding the bias term *b1(k)*. The ReLU activation function is then applied to the sum. The resulting tensor *fc1* has the shape *(batch_size, 1, 1, chan_dim)*.

Similarly, *fc2* represents the output of the second dense layer (Eq. 3). It further reduces the dimensionality of the input by multiplying the features from the first dense layer *fc1(i,1,1,d)* with the corresponding weights *W2(k,d)*, summing them up, and adding the bias term *b2(k)*. The sigmoid activation function is then applied to the sum. The resulting tensor *fc2* has the shape *(batch_size, 1, 1, chan_dim)*.

$$fc2(i,1,1,k) \, sig(W2(e,d).fc1(i,1,1,d+b2(k)) \quad (3)$$

In Eqs. (2) and (3), *e* refers to the index of the output feature maps in the dense layer. It ranges from *1* to the number of output channels (*chan_dim*); *d* refers to the index of the input feature maps in the dense layer. It ranges from *1* to the number of input channels (*C*).

The attention feature, *a_f*, is computed as,

$$a\_f(i,j,k) = fc2(i,1,1,k) * i\_f(i,j,k) \quad (4)$$

Eq. 4 computes the element-wise multiplication between the features received from the second dense layer *fc2(i,1,1,k)* and the input features *input_features(i,j,k)i.e, (i_f)*. This operation applies attention weights obtained from the second dense layer to each element of the input feature map.

Finally, the attention feature tensor (*a_f*) is added element-wise to the original input feature map, *i_f*, and is given as:

$$a\_f(i,j,k) = a\_f(i,j,k) + i\_f(i,j,k). \quad (5)$$

**Fine-tuning the Proposed Model:**

As stated earlier, the proposed methodology employs transfer learning by utilizing pretrained VGG19 model as feature extractor. The feature attention module is introduced to emphasize important channels within the extracted features. The modified features are then processed through dense layers for classification. By fine-tuning the pretrained model specifically for the task at hand, the model benefits from both the general visual representations learned from pretraining and the specialized attention mechanism.

IV. EXPERIMENTAL SETUP

As stated, experiment with the proposed neural network model is conducted using APTOS dataset. Details of this experimental setup have been described below.

**(A) Dataset Used:**

The APTOS dataset consists of a large collection of high-resolution retinal fundus images along with corresponding diagnostic labels provided by expert ophthalmologists. Table 1 shows the number of images used for different categories. There are five classes in the image dataset: Mild, Moderate, No DR, Proliferate, and Severe. A total of 6034 images were taken from 5 different classes. A sample image from each of the classes has been shown in Fig. 4.

**(B) Image Preprocessing:**

Each image is resized to 224x224 and normalized by dividing the pixel values by 255.

**(C) Performance Metrics Considered:**

The models' performance is evaluated using metrics such as accuracy, precision, recall, F1-score, Top1 % error and loss values. Confusion matrix is also considered.

**Table 1:** Images taken from APTOS 2019 dataset

| Types of Classes | Number of Images |
|---|---|
| Mild | 1624 |
| Moderate | 999 |
| No DR | 1805 |
| Proliferate | 772 |
| Severe | 834 |
| **Total Images** | **6034** |

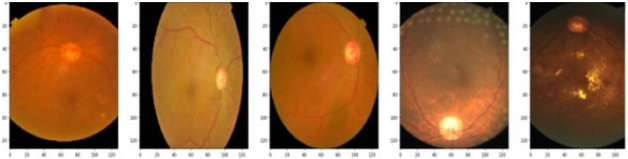

| (a) | (b) | (c) | (d) | (e) |

**Fig. 4:** Images taken from APTOS dataset. (a) Mild, (b) Moderate, (c) No DR, (d) Proliferate, and (e) Severe

**(D) Parameters Taken:**

Table 2 shows the parameters and their corresponding values used for our experimentation.

**TABLE 2:** Experimental setup

| Parameters | Values |
|---|---|
| Learning Rate | 0.0001 |
| Batch Size | 16 |
| Max Epochs | 40 |
| Optimizer | Adam |
| Loss Function | Categorical Cross-entropy |

**(E) Model Training:**

The dataset is split into training and test sets. The split is performed with a test size of 20% and stratified sampling to maintain class balance. During training, the models' weights are updated using backpropagation and gradient descent to minimize the loss function.

## V. ANALYSIS OF RESULTS

To evaluate the effectiveness of the incorporation of the feature attention module, APTOS data is considered. Experimentation was done on NVIDIA A100 tensor core GPU. A total of 20 simulations have been performed and the average scores are depicted into Table. 3. From the table it is noticed that the results are promising in nature in terms of various performance indices, yielding 96% (rounded) accuracy for enhanced VGG19 (with Feature Attention Block, denoted as, *FAB)*. Fig. 5 shows the variation in training and validation accuracy of VGG19 and VGG19+FAB. These curves indicate better performance when we add the new feature attention module with the pretrained model.

Likewise, Fig. 6 shows the variation in training and validation losses for VGG19 and VGG19+FAB. Fig. 6 confirms that introduction of the proposed feature attention module provides faster convergence with less loss.

**TABLE 3:** Results obtained using APTOS 2019 dataset

| Metrics | VGG19 + FAB (rounded) |
|---|---|
| Precision | 0.96 |
| Recall | 0.96 |
| F1-score | 0.96 |
| Accuracy (%) | 96 |
| Top-1 error (%) | 4.0 |

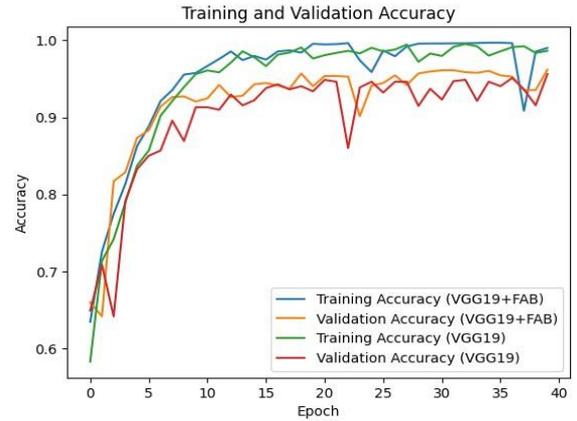

**Fig. 5:** Performance (accuracy curves) comparison between VGG19 and VGG19+FAB

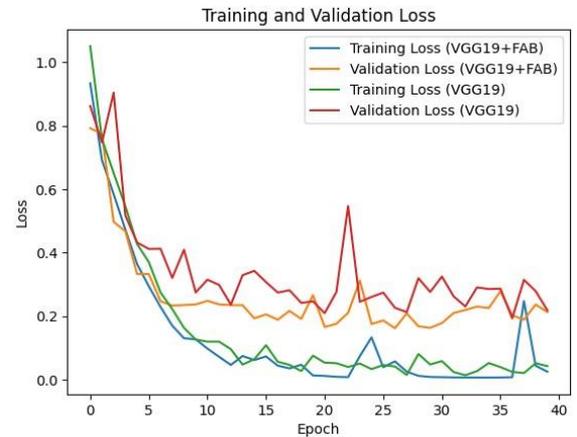

**Fig. 6:** Performance (loss curves) comparison between VGG19 and VGG19+FAB

**TABLE 4:** Accuracy values obtained with and without FAB for VGG19

| Neural Network Model | Accuracy (%) |
|---|---|
| VGG19 | 94.80 |
| VGG19 +FAB | **95.70** |

The accuracy values obtained with/without FAB are depicted in Table 4. This table establishes that incorporation of the feature attention module has an edge over the baseline VGG19 achieving an accuracy of 95.70%.

For visual illustration, predictions made by our fine-tuned model are shown in Fig 7 for four different classes of sample images. The figures corroborate our earlier findings on the superiority of the proposed net.

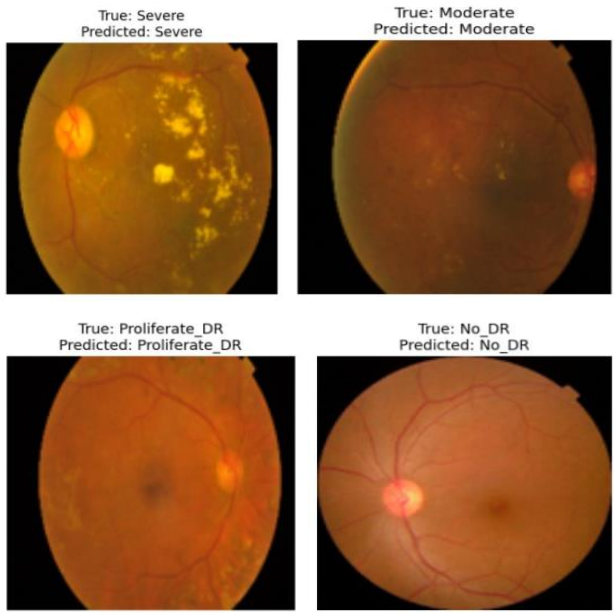

**Fig. 7:** Class predictions on unseen images by our proposed method (with the best results obtained using VGG19 + FAB)

The confusion matrix obtained for five different stages of DR using the proposed model (Fig. 8) indicates its efficacy for DR detection.

As stated, performance of the proposed model has also been compared with seven other state-of-the-art methods and the corresponding accuracy values are shown in Table 5. This table confirms the superiority of our proposed neural net model, augmented with attention block, for DR detection. Overall, the results demonstrate the efficacy of the incorporation of a feature attention module showcasing its potential in aiding early detection and diagnosis of DR.

**TABLE 5:** Classification performance of the proposed framework and the state-of-the-art models for the APTOS, 2019.

| Methods | Accuracy (%) |
| --- | --- |
| (Dondeti et al., 2020) [14] | 77.90 |
| (Bodapati et al., 2020) [15] | 81.70 |
| (Liu et al., 2020) [16] | 86.34 |
| (Kassani et al., 2019) [17] | 83.09 |
| (Bodapati et al., 2021) [18] | 82.54 |
| (Sikder et al., 2021) [19] | 94.20 |
| (Alyoubi et al., 2021) [20] | 89.00 |
| Proposed Framework | **95.70** |

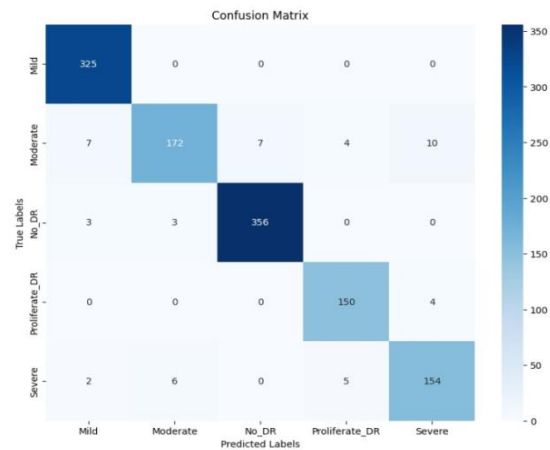

**Fig. 8:** Confusion matrix for five different stages of DR for the proposed method (with the best results obtained using VGG19 + FAB)

## V. CONCLUSION

The present work introduces a feature attention module in CNN for diabetic retinopathy detection. By fine-tuning the integrated feature attention block with a pretrained VGG19 model, we have achieved improved accuracy in identifying the severity levels of DR. The methodology leverages transfer learning from the VGG19. Additionally, the introduction of the channel attention block allows the model to selectively emphasize important channels, enhancing its ability to identify relevant features. Through experimentation and evaluation on the APTOS dataset, our proposed methodology demonstrated superior performance compared to the standalone VGG19 model and state-of-the-art methods confirming that the integration of the feature attention module could lead to enhanced discriminative power and improved accuracy in diabetic retinopathy detection.

Future work may explore employing the proposed methodology on larger and more diverse datasets.

## Acknowledgement

A part of this work has been supported by the IDEAS - Institute of Data Engineering, Analytics and Science Foundation, The Technology Innovation Hub at the Indian Statistical Institute, Kolkata through sanctioning a Project No /ISI/TIH/2022/55/ dtd. September 13, 2022.

## References

[1] Haykin, S. (2008). Neural Networks and Learning Machines. Pearson.
[2] Choudhury, D., Naug, A., & Ghosh, S. (2015, December). Texture and color feature based WLS framework aided skin cancer classification using MSVM and ELM. In 2015 Annual IEEE india conference (INDICON) (pp. 1-6). IEEE.
[3] Ross, T. J. (2004). Fuzzy Logic with Engineering Applications. Wiley.
[4] Dehuri, S., Ghosh, S., & Cho, S. (2011). Integration of swarm intelligence and artificial neural network. World Scientific, 78.
[5] Ghosh, S., & Chatterjee, A. (2023). Automated COVID-19 CT Image Classification using Multi-head Channel Attention in Deep CNN. arXiv preprint arXiv:2308.00715.

[6]  URL: https://www.researchgate.net/figure/General-Deep-Learning-Neural-Network_fig1_336179118.
[7]  Mello, V., Silva, G. F., Mendonça, J. S., & Lins, R. D. (2019). APTOS 2019 Blindness Detection. Kaggle. Available from: https://www.kaggle.com/c/aptos2019-blindness-detection.
[8]  Gulshan, V., Peng, L., Coram, M., Stumpe, M. C., Wu, D., Narayanaswamy, A., ... & Webster, D. R. (2016). Development and validation of a deep learning algorithm for detection of diabetic retinopathy in retinal fundus photographs. JAMA, 316(22), 2402-2410.
[9]  Abràmoff, M. D., Lou, Y., Erginay, A., Clarida, W., Amelon, R., Folk, J. C., & Niemeijer, M. (2016). Improved automated detection of diabetic retinopathy on a publicly available dataset through integration of deep learning. Investigative Ophthalmology & Visual Science, 57(13), 5200-5206.
[10] Quellec, G., Charrière, K., Boudi, Y., Cochener, B., & Lamard, M. (2017). Deep image mining for diabetic retinopathy screening. Medical Image Analysis, 39, 178-193.
[11] Burlina, P. M., Joshi, N., Pekala, M., Pacheco, K. D., Freund, D. E., & Bressler, N. M. (2017). Automated grading of age-related macular degeneration from color fundus images using deep convolutional neural networks. JAMA Ophthalmology, 135(11), 1170-1176.
[12] Chaudhuri, S., Biyani, P., & Singh, S. (2018). A comprehensive analysis of deep learning models for diabetic retinopathy detection. Journal of Healthcare Engineering, 2018, 2948769.
[13] Gargeya, R., & Leng, T. (2017). Automated identification of diabetic retinopathy using deep learning. Ophthalmology, 124(7), 962-969.
[14] Dondeti, V., Bodapati, J.D., Shareef, S.N., & Veeranjaneyulu, N. (2020). Deep convolution features in non-linear embedding space for fundus image classification. Revue d'Intelligence Artificielle, 34(3), 307-313.
[15] Bodapati, J.D., Naralasetti, V., Shareef, S.N., Hakak, S., Bilal, M., Maddikunta, P.K.R., & Jo, O. (2020). Blended multi-modal deep convnet features for diabetic retinopathy severity prediction. Electronics, 9(6), 914.
[16] Liu, H., Yue, K., Cheng, S., Pan, C., Sun, J., & Li, W. (2020). Hybrid model structure for diabetic retinopathy classification. Journal of Healthcare Engineering, 2020.
[17] Kassani, S.H., Kassani, P.H., Khazaeinezhad, R., Wesolowski, M.J., Schneider, K.A., & Deters, R. (2019). Diabetic retinopathy classification using a modified Xception architecture. In 2019 IEEE International Symposium on Signal Processing and Information Technology (ISSPIT) (pp. 1-6). IEEE.
[18] Bodapati, J.D., Shaik, N.S., & Naralasetti, V. (2021). Composite deep neural network with gated-attention mechanism for diabetic retinopathy severity classification. Journal of Ambient Intelligence and Humanized Computing, 1-15.
[19] Sikder, N., Masud, M., Bairagi, A.K., Arif, A.S.M., Nahid, A.A., & Alhumyani, H.A. (2021). Severity classification of diabetic retinopathy using an ensemble learning algorithm through analyzing retinal images. Symmetry, 13(4), 670.
[20] Alyoubi, W.L., Abulkhair, M.F., & Shalash, W.M. (2021). Diabetic retinopathy fundus image classification and lesions localization system using deep learning. Sensors, 21(11), 3704.